\documentclass[journal,twoside,web]{ieeecolor2}
\usepackage{generic}
\usepackage{cite}
\usepackage{amsmath,amssymb,amsfonts}
\usepackage{algorithmic}
\usepackage{graphicx}
\usepackage[outline]{contour}
\usepackage[pdfstartview = XYZ, pdftex, hyperindex = false,  bookmarks = false, colorlinks = true, linkcolor = black, urlcolor = black, citecolor = black]{hyperref}
\usepackage{cite}
\usepackage{booktabs}
\usepackage{multirow}
\usepackage{microtype}
\usepackage{adjustbox}
\usepackage{colortbl}
\usepackage{array}
\usepackage[dvipsnames]{xcolor}
\usepackage{orcidlink}

\newcolumntype{C}[1]{>{\centering\arraybackslash}p{#1}}

\newcommand{\bolduparrow}{\contourlength{0.03em}\contour{black}{\ensuremath{\uparrow}}}
\newcommand{\bolddownarrow}{\contourlength{0.03em}\contour{black}{\ensuremath{\downarrow}}}

\newcommand{\etal}{\textit{et al.}}
\newcommand{\eg}{\textit{e.g.,}}

\usepackage{microtype}
\usepackage{textcomp}
\def\BibTeX{{\rm B\kern-.05em{\sc i\kern-.025em b}\kern-.08em
    T\kern-.1667em\lower.7ex\hbox{E}\kern-.125emX}}
\begin{document}
\title{Surrogate Supervision for Robust and Generalizable Deformable Image Registration}
\author{Yihao~Liu,$^{\orcidlink{0000-0003-3187-9903}}$\IEEEmembership{Member, IEEE}, Junyu~Chen,$^{\orcidlink{0000-0003-4672-6408}}$\IEEEmembership{Member,~IEEE}, Lianrui~Zuo,$^{\orcidlink{0000-0002-5923-9097}}$\IEEEmembership{Member,~IEEE},  Shuwen~Wei,$^{\orcidlink{0000-0001-8679-9615}}$\IEEEmembership{Member,~IEEE},  Brian~D.~Boyd, Carmen~Andreescu, Olusola~Ajilore, Warren~D.~Taylor,$^{\orcidlink{0000-0002-9975-3082}}$ Aaron~Carass,$^{\orcidlink{0000-0003-4939-5085}}$\IEEEmembership{Member,~IEEE},~and~Bennett~A.~Landman,$^{\orcidlink{0000-0001-5733-2127}}$\IEEEmembership{Fellow,~IEEE}
\thanks{This paragraph of the first footnote will contain the date on 
which you submitted your paper for review.}
\thanks{
This research is supported by the National Artificial Intelligence Research Resource~(NAIRR) Pilot allocation NAIRR240482 and the Jetstream2 cloud resource supported by the National Science Foundation (award NSF-OAC 2005506) at Indiana University.
}
\thanks{Y. Liu and L. Zuo is with the Department of Electrical and Computer Engineering, Vanderbilt University, Nashville, TN, USA. Y.~Liu~(e-mail: yihao.liu\@vanderbilt.edu) is supported by NIH grants R01HL169944, U24AG074855, and R01MH121620.}
\thanks{J. Chen is with the Department of Radiology and Radiological Science, Johns Hopkins Medical School, Baltimore, MD, USA.}
\thanks{S. Wei and A. Carass are with the Image Analysis and Communications Laboratory in the Department of Electrical and Computer Engineering, Johns Hopkins University, Baltimore, MD, USA. Both are supported in part by NIH grants U01-NS122764, U01-NS112120, U01-NS111678, R01-EY032284, and R01-EB036013, as well as JH Discovery Grants.}
\thanks{B.~D.~Boyd is with the Center for Cognitive Medicine, Department of Psychiatry and Behavioral Science, Vanderbilt University Medical Center, Nashville, TN, USA.}
\thanks{C.~Andreescu, MD, Professor of Psychiatry and Bioengineering, is with the University of Pittsburgh, School of Medicine, PA, USA. C.~Andreescu is supported by in part by NIH grants R01-MH121619~(Rembrandt) and R01-MH108509.}
\thanks{O.~Ajilore, MD, PhD, is with the Department of Psychiatry, University of Illinois College of Medicine, IL, USA. Dr.~Ajilore is funded in part by NIMH~R01MH121384. Dr.~Ajilore is the co-founder of KeyWise AI. He is also a consultant for Otsuka Pharmaceutical and serves on the advisory boards of Embodied Labs and Blueprint.}
\thanks{W.~D.~Taylor is with the Center for Cognitive Medicine, Department of Psychiatry and Behavioral Science, Vanderbilt University Medical Center, Nashville, TN, USA and the Geriatric Research, Education, and Clinical Center, Veterans Affairs Tennessee Valley Health System, Nashville, TN, USA. W.~D.~Taylor is supported in part by NIH grant R01-MH121620}
\thanks{B.~A.~Landman is with the Departments of Computer Science, Electrical Engineering, and Biomedical Engineering, Vanderbilt University, USA and the Department of Radiology and Radiological Sciences, Vanderbilt University Medical Center, USA.}
}

\maketitle

\begin{abstract}

\textit{Objective}:
Deep learning-based deformable image registration has achieved strong accuracy, but remains sensitive to variations in input image characteristics such as artifacts, field-of-view mismatch, or modality difference.
We aim to develop a general training paradigm that improves the robustness and generalizability of registration networks.
\textit{Methods}: 
We introduce surrogate supervision, which decouples the input domain from the supervision domain by applying estimated spatial transformations to surrogate images.
This allows training on heterogeneous inputs while ensuring supervision is computed in domains where similarity is well defined.
We evaluate the framework through three representative applications: artifact-robust brain MR registration, mask-agnostic lung CT registration, and multi-modal MR registration.
\textit{Results}:
Across tasks, surrogate supervision demonstrated strong resilience to input variations including inhomogeneity field, inconsistent field-of-view, and modality differences, while maintaining high performance on well-curated data.
\textit{Conclusions}:
Surrogate supervision provides a principled framework for training robust and generalizable deep learning-based registration models without increasing complexity.
\textit{Significance}:
Surrogate supervision offers a practical pathway to more robust and generalizable medical image registration, enabling broader applicability in diverse biomedical imaging scenarios.
\end{abstract}


\section{Introduction}
\label{sec:introduction}
\IEEEPARstart{D}{eformable} image registration plays a central role in medical image analysis by enabling the alignment of anatomical structures across images and subjects.
Deep learning-based approaches have recently shown impressive performance in this domain, offering accurate and efficient estimation of spatial transformations~\cite{chen2025survey}.
As these methods gain traction, there is an increasing interest in developing models that not only have good performance in well-curated data, but are also robust to variations in imaging modality, contrast, resolution, and artifacts---conditions that are common in real-world medical images.
Addressing such variability is critical for building registration models that generalize across datasets, scanners, and institutions.
Most existing approaches are developed on data that have been preprocessed using specific pipelines to ensure consistency in input characteristics~\cite{liu2022coordinate,tian2024unigradicon,balakrishnan2019tmi,hoffmann2021tmi}.
Such approaches also ensure that the similarity loss between the fixed and moving images can be computed in a reliable manner.
However, this reliance often leaves models tightly coupled to the particular preprocessing procedures used during training.
Achieving similar performance in real-world settings requires users to replicate the same preprocessing steps.
Others attempt to improve generalization by scaling up the training set, incorporating data from multiple sources, modality, or even anatomical structures~\cite{tian2024unigradicon}.
While this increases data diversity, it places the burden on the model to implicitly learn an invariance to factors such as artifacts, field-of-view mismatch, or modality difference, where direct computation of similarity between inputs can be unreliable.

Traditionally, deep learning based registration methods rely on direct similarity losses between the input image pairs to provide supervision.
However, one insight that has emerged in prior work is that supervision in training registration models does not need to be restricted to these direct comparisons.
In particular, VoxelMorph~\cite{balakrishnan2019tmi} and SynthMorph~\cite{hoffmann2021tmi} apply the predicted deformation field to anatomical label maps and supervise the training by comparing the warped and fixed label maps.
This setup enables training using label overlap metrics, which offer an additional source of supervision beyond intensity similarity.
Cao~\etal~developed a multi-modal registration model that is supervised by a mono-modal similarity measure using available paired data~\cite{cao2018deep}.
In this work, we propose a generalized framework that unifies and extends these prior methods,, which we term \textbf{surrogate supervision}.

The core concept of surrogate supervision is to decouple the model's input domain from the domain in which supervision is applied.
A registration model estimates a spatial transformation between two images, which is typically applied to the moving image to create a warped image.
The similarity loss used in training is computed by comparing the warped image with the fixed image.
However, the estimated transformation does not have to be applied to the exact input image for the registration network to provide supervision.
Instead, it can be applied to any surrogate of the moving image, such as a preprocessed version, a scan with different imaging modality acquired from the same subject, or a synthetically augmented variant. The similarity loss can then be computed between the warped moving surrogate and a corresponding surrogate of the fixed image.
We are free to select surrogate images where similarity is well-defined and reliable, even when the original input images are heterogeneous.
With this formulation, we can leverage not only surrogates such as paired multi-modal data or anatomical label maps---as done in prior work---but also incorporate the knowledge encoded in various established preprocessing pipelines, such as artifact removal or region-of-interest extraction, into the training process.
Surrogate supervision provides a principled mechanism for integrating these diverse sources of domain knowledge to improve the robustness and generalizability without adding complexity at inference time.

In this work, we explored two representative applications: (1)~learning artifact robust registration, and (2)~learning mask agnostic registration.
We further include a multi-modal registration experiment to examine how surrogate supervision compares with contemporary multi-modal similarity loss functions for multi-modal registration.
These applications of surrogate supervision are not exhaustive, but serve to demonstrate the broader applicability and inspire new directions for developing adaptable and generalizable registration models.

\section{Related Works}
\label{sec:related_works}
\subsection{Deep Learning Registration}
While traditional optimization-based methods remain effective for rigid and affine registration, recent studies suggest that deep learning-based deformable registration methods have the potential to achieve competitive or superior performance in terms of both accuracy~\cite{liu2022coordinate, tian2024unigradicon, chen2022mia, tan2024tmi} and generalizability~\cite{tian2024unigradicon, chen2025beyond, hering2023tmi}.
These methods leverage neural network models to learn a function that can estimate spatial transformations in a single forward pass, and they form the basis for the work presented in this paper.

Most learning-based deformable registration models are trained in an unsupervised, or more accurately, self-supervised manner.
In this setup, the model predicts a deformation field, which is then applied to the moving image to generate a warped image.
The loss function used during training typically takes the form of
\begin{equation}
    L(I_f, I_m, \phi) = \mathcal{L}_{\text{Sim}}(I_f, I_m\circ\phi) + \lambda\mathcal{L}_{\text{Reg}}(\phi),
    \label{e:energy}
\end{equation}
where the term $\mathcal{L}_{\text{Sim}}$ penalizes intensity dissimilarities between the fixed image $I_f$ and the warped image $I_m\circ\phi$, and the regularization term $\mathcal{L}_{\text{Reg}}$ encourages the predicted deformation field $\phi$ to be smooth, with $\lambda$ used to balance these terms.

The similarity term $\mathcal{L}_{\text{Sim}}$ plays a central role in guiding the optimization.
In mono-modal registration, where the corresponding structures across images are expected to have consistent intensity profiles, similarity losses such as mean squared error~(MSE) and normalized cross-correlation~(NCC) are commonly used~\cite{chen2025beyond, chen2025survey}. Multi-modal registration, however, may require more sophisticated strategies, as discussed below.

\subsection{Multi-modal Registration Losses}
Multi-modal registration presents a greater challenge, as the same anatomical structures may appear with very different intensities or be absent entirely.
In traditional optimization-based registration, mutual information~\cite{collignon1995ipmi, wells1996mia} has been a widely used and successful similarity measure.
Before mutual information and related losses were incorporated into deep learning–based registration models, researchers addressed the multi-modal registration problem using paired multi-modal data or by incorporating image synthesis.
Specifically, Cao~\etal~\cite{cao2018deep} leverage pre-aligned CT–MR training pairs to transform the difficult task of defining a multi-modal similarity measure into a simpler mono-modal problem.
During training, each unaligned CT–MR pair is supervised by comparing the warped moving image to the fixed image’s paired counterpart in the same modality, enabling the network to learn multi-modal alignment indirectly through reliable mono-modality similarity measures.
Alternatively, Arar~\etal~\cite{arar2020unsupervised} avoid the need for paired data by employing a geometry-preserving image-to-image translation network to synthesize one modality from the other while maintaining structural consistency, followed by intra-modality registration between the synthesized and target images.
More recently, similarity losses such as mutual information~(MI)~\cite{guo2019multi}, correlation ratio~\cite{roche1998miccai,chen2025correlation}, and others~\cite{heinrich2012mia, chen2015spie, chen2017mia} have been explored. However, in deep learning contexts, MI and correlation ratio require approximating intensity statistics with continuous functions (e.g., Parzen windowing) for auto-differentiation, which introduces a tradeoff between bin resolution and computational burden.

\begin{figure*}
    \centering
    \includegraphics[width=0.8\linewidth]{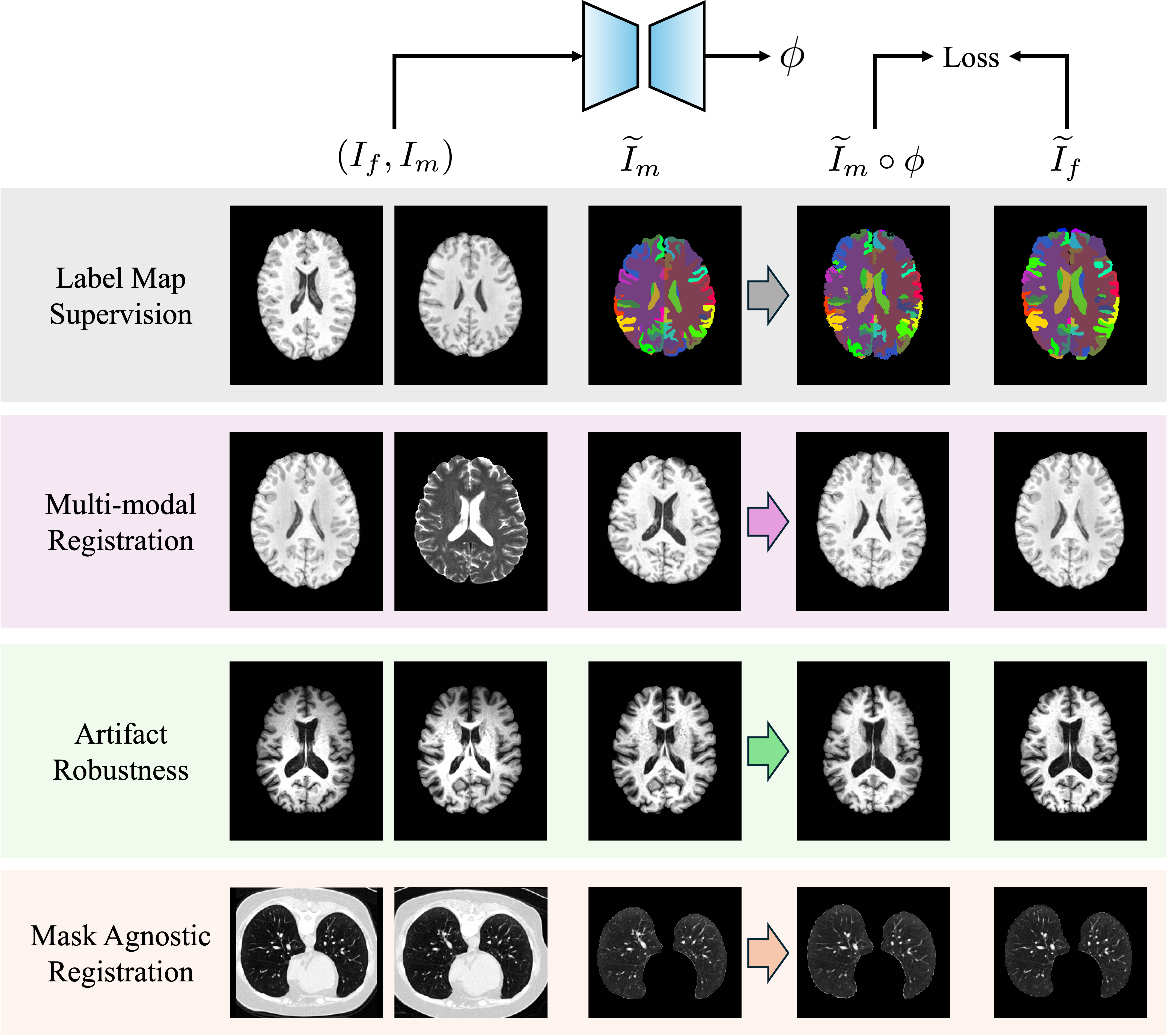}
    \caption{Surrogate supervision framework and its applications in label map supervision, multi-modal registration, artifact robustness, and mask agnostic registration. Although the inputs to the registration model may be heterogeneous, we always enforce supervision in a homogeneous domain.\label{f:method}}
\end{figure*}

\subsection{Label Map Supervision}
Beyond intensity-based losses, anatomical label maps have also been widely employed as an additional source of supervision for deep learning models.
Methods like VoxelMorph~\cite{balakrishnan2019tmi} and SynthMorph~\cite{hoffmann2021tmi} apply the predicted deformation field to external label maps and train the model by comparing the warped labels to fixed labels. This process is shown in Figure~\ref{f:method}(a).
In this setting, the Dice loss is commonly used to encourage label overlap.
Label map supervision provides a structure-aware training signal that complements intensity-based objectives and can (potentially) help the model better capture deformations.

\subsection{Towards Generalizable Model} With the development of deep learning, there has been growing interest in building robust and generalizable medical AI models that can perform effectively across diverse imaging modalities, scanners, and clinical environments.
This includes emerging efforts toward building foundation models for registration.
For example, Tian~\textit{et al.}~\cite{tian2024unigradicon} utilizes large-scale datasets to improve generalization. Li~\etal~\cite{li2025unireg} introduces a conditional control vector to encode anatomical part, registration type, and instance-specific features, enabling task-specific deformation field generation.
Chen~\textit{et al.}~\cite{chen2025midl} and Dey~\textit{et al.}~\cite{dey2024learning} both employ synthetic data generation to expose the model to a wide range of imaging variations.
These approaches address data variability primarily through expanding the diversity of training samples.

\section{Method}
\label{sec:method}
Let $I_f$ denote the fixed image and $I_m$ denote the moving image.
Deep learning-based deformable image registration algorithms estimate a spatial transformation $\phi$ that aligns $I_m$ to $I_f$ using a neural network model $N_\theta$, parameterized by $\theta$:
\begin{equation}
    \phi = N_\theta(I_f, I_m).
    \label{e:network}
\end{equation}
Training involves minimizing a self-supervised objective shown in Equation~\ref{e:energy} with respect to $\theta$.

\subsection{Surrogate Supervision: Framework}
While the deformation field $\phi$ is estimated from the original input pair $(I_f, I_m)$, we observe that the loss can be computed in a different domain that provides more reliable or meaningful supervision.
To enable this, we introduce surrogate images $(\widetilde{I}_f, \widetilde{I}_m)$ corresponding to the original input pair.
These surrogates may differ from the original images in modality, preprocessing, or other properties.
Examples of such surrogates are provided in Figure~\ref{f:method}.
While the model $N_\theta$ still takes $(I_f, I_m)$ as input, as defined in Equation~\ref{e:network}, the surrogate supervision objective is given by:
\begin{equation}
    L_s(\widetilde{I}_f, \widetilde{I}_m, \phi) = \mathcal{L}_{\text{Sim}}(\widetilde{I}_f, \widetilde{I}_m\circ\phi) + \lambda\mathcal{L}_{\text{Reg}}(\phi).
    \label{e:loss}
\end{equation}
This formulation decouples the model’s input domain $(I_f, I_m)$ from the supervision domain $(\widetilde{I}_f, \widetilde{I}_m)$, enabling flexible selection of the domain in which supervision is applied. Importantly, the deformation field is still predicted from the original input image pair, and the gradient signal from the loss computed on the surrogate domain propagates through the differentiable warping operation~\cite{jaderberg2015spatial} back to the model parameters.
This ensures that supervision in the surrogate domain effectively trains the model, even if the input and supervision domains differ.
% Intuitively, the network is trained on $(I_f, I_m)$ but supervised on $(\widetilde{I}_f, \widetilde{I}_m)$, which are not required at test time. 
% Through learning, the network implicitly links surrogates to the original inputs, allowing registration without access to surrogate images during inference, though instance-based optimization remains possible if they are available.
Notably, this strategy only modifies the training process. Once the model is trained, it can be applied directly to new image pairs without requiring surrogate images.

Surrogate supervision is compatible with a broad range of model architectures and loss functions, since it does \textit{not} impose constraints on the structure of $N_\theta$ or the form of $\mathcal{L}_{\text{Sim}}$.
We note that multi-modal registration through mono-modal supervision using paired data~\cite{cao2018deep} and label map supervision~\cite{balakrishnan2019tmi} described in Section~\ref{sec:related_works} can be viewed as two examples of this framework.
In the following, we demonstrate how alternative choices of surrogate domains~(beyond anatomical labels and pre-aligned multi-modal data) can be used to tackle challenges in registration, including improving artifact robustness, and learning mask agnostic registration.

\subsection{Surrogate Supervision: Applications}

\subsubsection{Artifact Robust Registration}
Medical images exhibit varying quality and are often affected by artifacts arising from acquisition protocols, scanner hardware, and reconstruction techniques.
These artifacts can impair the visibility of anatomical structures.
For example, MRI scans frequently suffer from intensity inhomogeneity introduced by coil sensitivity or magnetic field non-uniformities~\cite{smith2010im}; CT images may exhibit beam hardening and other artifacts~\cite{barrett2004rg}; diffusion MRI can be affected by eddy currents and susceptibility distortions~\cite{le2006jmri}; and ultrasound images often contain speckle noise and other artifacts~\cite{lacefield2014chapter}.
Although deep learning models can benefit from training on diverse datasets that reflect this variability, there remains a fundamental challenge at the level of supervision:
\textit{traditional similarity losses may not be meaningful, and constructing a dedicated loss is itself a challenging problem.}
In particular, comparing raw intensities affected by differing artifacts can lead to inaccurate gradients during training due to mismatched intensity profiles in the corresponding structures.

Traditionally, the medical imaging community has developed specialized algorithms to correct for such artifacts.
For example, N4 inhomogeneity correction~\cite{tustison2010tmi} is widely used to mitigate intensity inhomogeneity in MRI; metal artifact reduction techniques~\cite{glover1981mp, wang1996tmi} are employed in CT; eddy current correction~\cite{jezzard1998mrm} is applied in diffusion MRI; and speckle reduction~\cite{kremkau1986jum} is used in ultrasound imaging.
These algorithms have been extensively validated and are routinely used in clinical and research workflows.

Rather than attempting to relearn these algorithms from scratch or requiring the user to preprocess the data, surrogate supervision provides a principled way to leverage existing, well-validated techniques during model training.
Consider training a model to register MR images affected by intensity inhomogeneity, as shown in Figure~\ref{f:method}(c).
Two raw MR images, $I_f$ and $I_m$, are provided as input to the network.
The predicted deformation field $\phi$ is then applied to the bias-corrected moving surrogate $\widetilde{I}_m$.
The similarity loss is computed between $\widetilde{I}_m$ and the fixed image's surrogate $\widetilde{I}_f$, which in this instance would be the bias-corrected fixed image.
Because both surrogates have been corrected for inhomogeneity, standard similarity losses can be reliably applied, as we would expect the intensity profiles at the corresponding anatomical structures to be consistent.

Through surrogate supervision, the model learns to extract features that are inherently robust to intensity inhomogeneity from inputs with intensity inhomogeneity.
Once the model is trained, it requires no additional input beyond the raw images $I_f$ and $I_m$ at inference time.
This flexibility enhances the generalizability of the trained model, making it suitable for a wide range of real-world use cases without the requirement of artifact correction preprocessing at deployment time.

\subsubsection{Mask Agnostic Registration}
It is common in medical image processing workflows to apply masking to an image to focus the analysis on specific regions of interest~(ROIs) and reduce irrelevant variability.
For example, brain MRI scans are often skull-stripped to isolate intracranial structures, while thoracic CT scans frequently undergo lung extraction to target pulmonary anatomy.
Optimization-based registration tools such as the Advanced Normalization Tools~(ANTs)~\cite{avants2011reproducible, avants2009ij} support optional mask inputs to restrict optimization to the specified ROI for each pair of inputs.
In contrast, deep learning methods typically require a design decision at training time: whether to use masked or unmasked images.
This choice dictates how the model learns to interpret the image content and implicitly assumes that the same masking strategy will be applied during deployment.
This coupling between training and deployment assumptions can limit model flexibility and generalizability.
One possible remedy is to train the model with a mixture of masked and unmasked inputs. Yet this forces the network to handle multiple tasks simultaneously, which can weaken its performance in any single setting. It may also fail to address cases where the fixed and moving images are masked differently and thus contain mismatched anatomical content or fields of view, making direct similarity losses inherently unreliable.

With surrogate supervision, the model may take either masked or unmasked images as input, while the similarity loss is computed using masked versions to ensure that optimization focuses on the specified region of interest.
A visual example is provided in Figure~\ref{f:method}(d).
As a result, the trained model does not require explicit masks at inference time and remains robust regardless of whether the input images are pre-masked.

\begin{figure*}
    \centering
    \begin{tabular}{ccc}
        \includegraphics[width=0.3\linewidth]{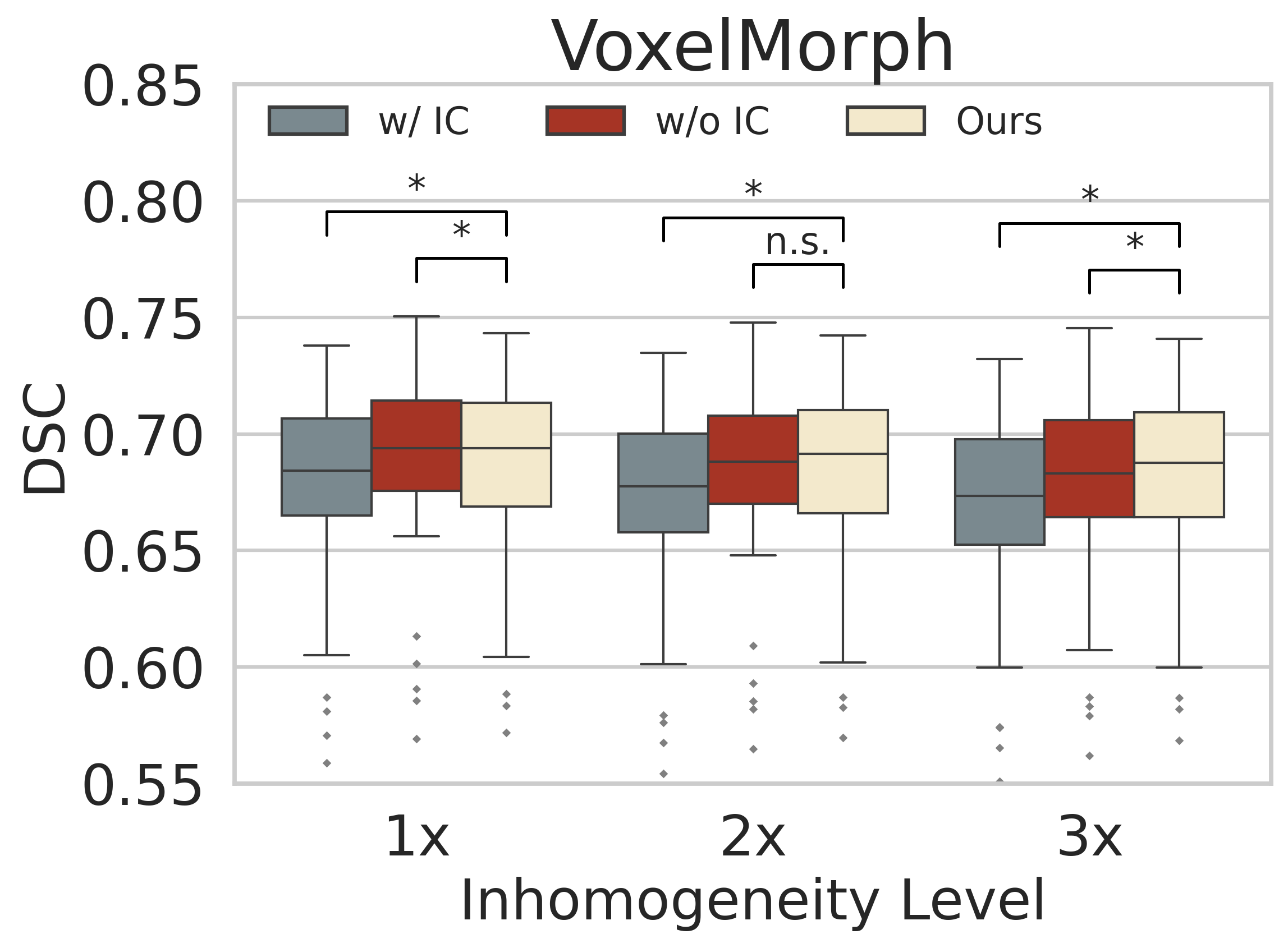}
        &
        \includegraphics[width=0.3\linewidth]{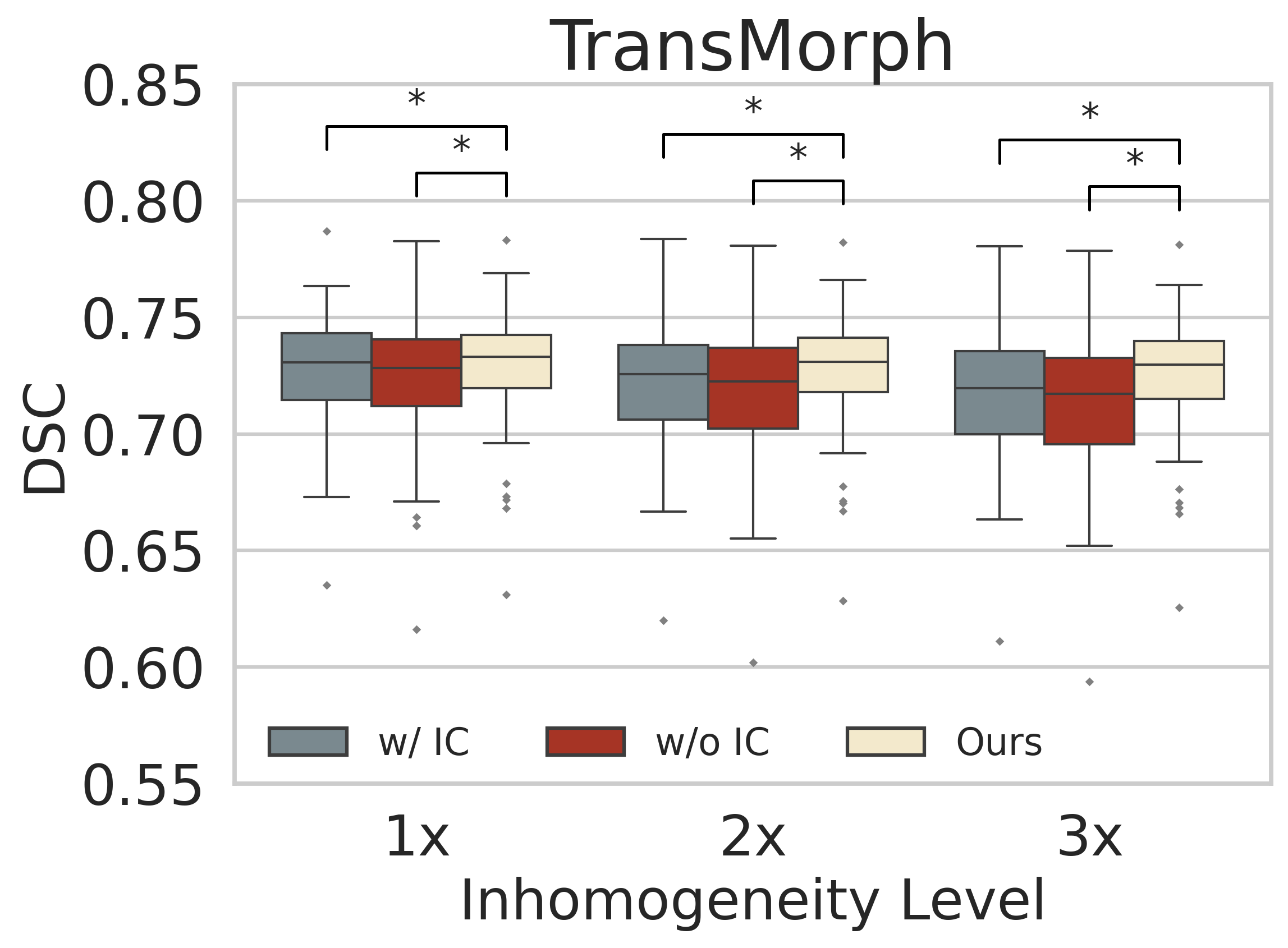}
        &
        \includegraphics[width=0.3\linewidth]{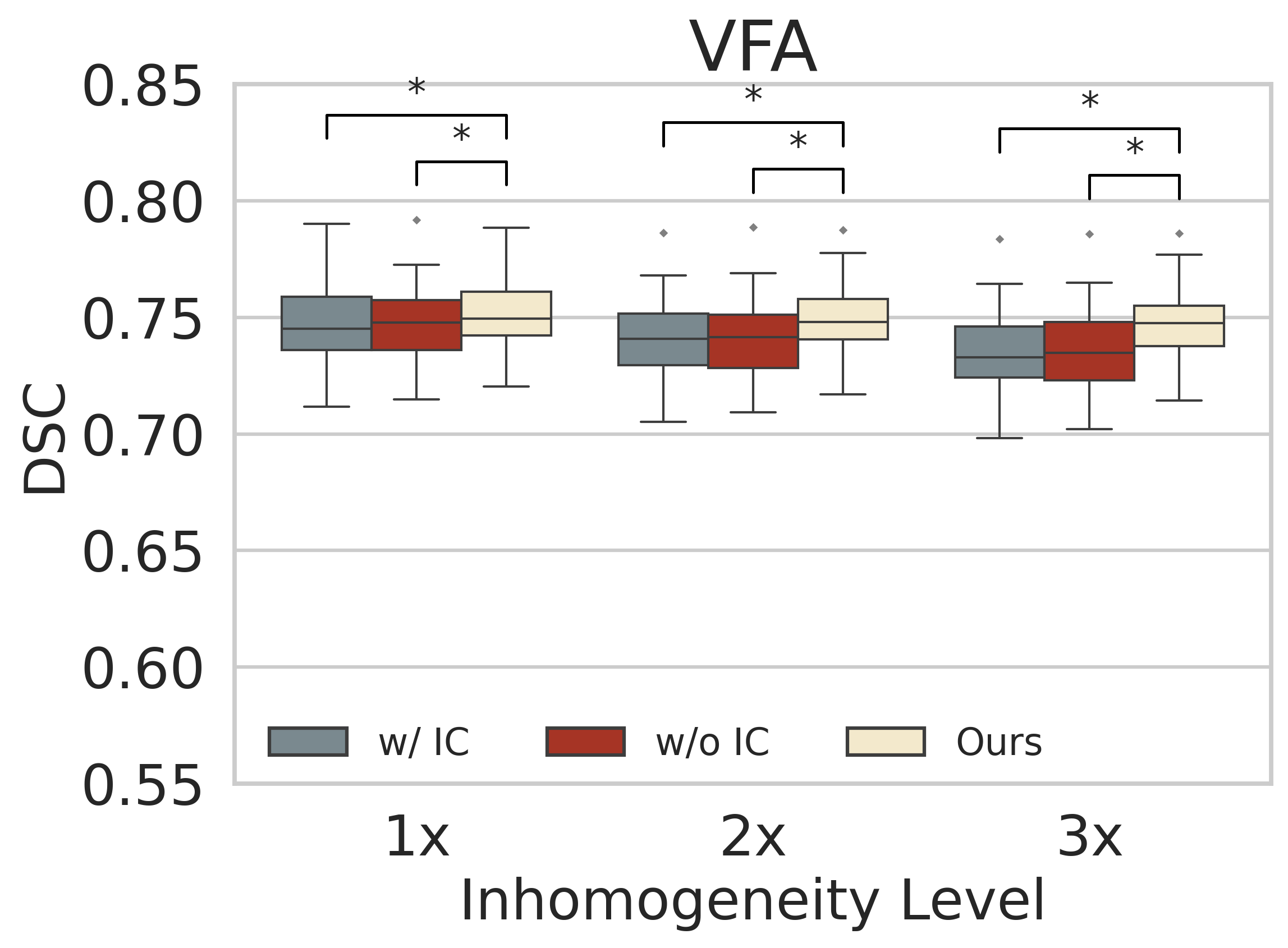}
    \end{tabular}
    
    \caption{Quantitative evaluation of registration robustness to MR intensity inhomogeneity. We report the mean Dice Similarity Coefficient~(DSC) for each architecture (VoxelMorph, TransMorph, VFA) under three training configurations: with inhomogeneity-correction~(w/~IC), without inhomogeneity-corrected~(w/o~IC), and surrogate supervision~(Ours). Separate box plots are shown for each architecture, with DSC measured at increasing inhomogeneity levels (1$\times$, 2$\times$, 3$\times$). Within each inhomogeneity level, we performed a paired Wilcoxon signed-rank test~\cite{wilcoxon1945bb} between surrogate supervision~(Ours) and the other two training configurations (w/~IC and w/o~IC). Brackets above the box plots indicate the results of these tests, with ``$*$'' denoting significance at an $\alpha$ of $0.01$ and  not significant is denoted with ``n.s.''.\label{f:ic_boxplot}}
\end{figure*}

\section{Experiments}
\subsection{Experimental Setup}
We evaluated the proposed surrogate supervision framework using three publicly available registration models: VoxelMorph~\cite{balakrishnan2019tmi}, TransMorph~\cite{chen2022mia}, and VFA~\cite{liu2024jmi}.
Model implementations were obtained from their GitHub repositories.
All models were trained using the Adam optimizer with a learning rate of $1\times10^{-4}$ and a batch size of 1, following recommendations from the original papers.
Training continued until the validation loss failed to improve for $50,000$ steps, up to a maximum of $500,000$ steps.

We assessed registration accuracy using either the Dice similarity coefficient~(DSC) for label overlap or the target registration error~(TRE) for landmark alignment, depending on the availability of segmentation maps or keypoint annotations for each individual task.
To evaluate the regularity of the predicted transformations, we calculated the non-diffeomorphic volume~\cite{liu2022finite}.
For statistical analysis, we used the paired Wilcoxon signed-rank test~($\alpha=0.01$) under the null hypothesis that the distribution of differences between two methods is symmetric about zero.

\begin{figure*}
    \begin{tabular}{ccccc}
        Inhomogeneity field & $\times0$ & $\times1$ & $\times2$ & $\times3$
        \\
        \includegraphics[width=0.18\textwidth]{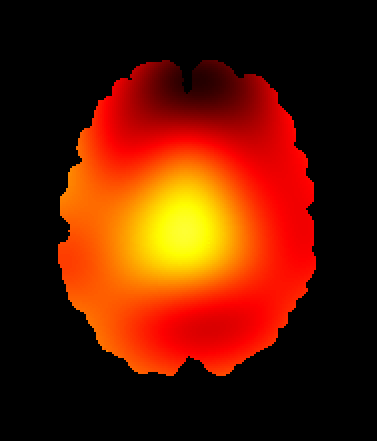}
        &
        \includegraphics[width=0.18\textwidth]{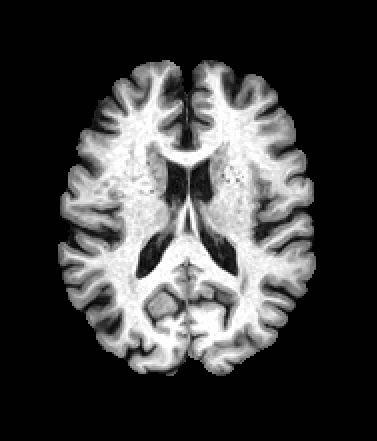}
        &
        \includegraphics[width=0.18\textwidth]{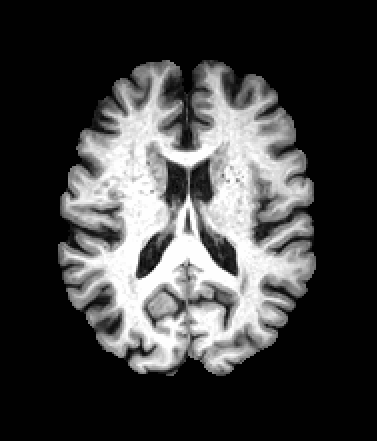}
        &
        \includegraphics[width=0.18\textwidth]{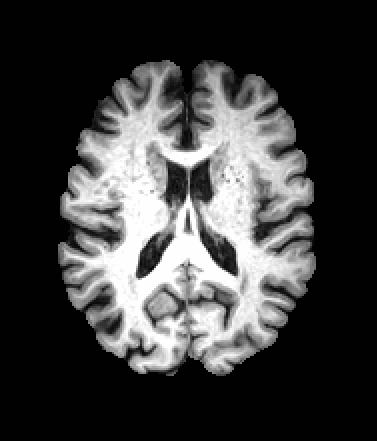}
        &
        \includegraphics[width=0.18\textwidth]{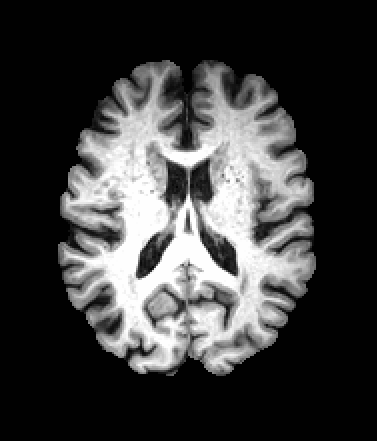}
    \end{tabular}
    \caption{Examples of test images used to assess artifact robustness. Different degrees of inhomogeneity~(up to $\times3$) were simulated using the N4-estimated inhomogeneity field. The corresponding inhomogeneity-corrected image~($\times0$) is shown for reference.
    \label{f:ic}
    }
\end{figure*}
\subsection{Artifact Robust Registration}
We used MR intensity inhomogeneity as an example to demonstrate the use of surrogate supervision in learning artifact robust registration. 
In this experiment, we used a proprietary dataset from the REMBRANDT study~\cite{taylor2024mad}, which comprises older participants aged 60 years and above.
The study was approved by the relevant institutional review boards, and all participants provided written informed consent.
The training set comprising $226$ imaging sessions from $68$ subjects.
The validation and test sets included $24$ and $48$ fixed pairs from 15 subjects each for inter-subject registration.
All scans were preprocessed with skull stripping via HD-BET~\cite{isensee2019hbm} and rigid alignment using ANTs~\cite{avants2009ij}.
Training involved selecting two random T1-weighted images to serve as the moving and fixed inputs, respectively.
Three models were trained per architecture: (1)~trained on inhomogeneity-corrected images~(w/~IC), (2)~trained on raw uncorrected images~(w/o~IC), and (3)~trained with surrogate supervision, using uncorrected inputs but computing supervision on inhomogeneity-corrected surrogates~(Ours).
All models used mean squared error as the similarity loss with $\lambda = 1$, and were evaluated using DSC with segmentation label maps generated by a whole-brain segmentation algorithm~\cite{yu2023unest}.
To simulate varying degrees of inhomogeneity, we scaled the N4-estimated inhomogeneity field strength up to 3$\times$ at test time.
Figure~\ref{f:ic} shows example test images with varying inhomogeneity levels.
Figure~\ref{f:ic_boxplot} reports the quantitative results.

For both the TransMorph and VFA models, surrogate supervision leads to significantly better performance compared to other training configurations. For VoxelMorph, surrogate supervision yield similar performance as training directly on uncorrected raw images for inhomogeneity levels 1$\times$ and 2$\times$. We attribute this to the limited capacity of the VoxelMorph architecture.
Since artifact robustness requires the model to learn a more complex mapping.
Simpler models like VoxelMorph may be underparameterized for this task, and thus unable to fully capitalize on the benefits of surrogate supervision.
We also observe that the performance of both w/~IC and w/o~IC models degrade as the inhomogeneity level increases.
In contrast, the surrogate-supervised models maintain a consistent performance.
We attribute this robustness to the use of both raw image as input and clean surrogate supervision: w/~IC models lack exposure to artifact variability for robustness, while w/o~IC models suffer from inferior supervision during training.

\begin{table*}[!ht]
    \centering
    \caption{Quantitative results for lung CT inhale-exhale registration on the NLST dataset. We report mean Target Registration Error~(TRE) in millimeters on expert-annotated landmarks within the lungs. Each architecture (VoxelMorph, TransMorph, VFA) is trained under four settings: Masked: both inputs masked by ROI; Unmasked: neither input masked; Random: both inputs either masked or unmasked, chosen randomly per sample; and Ours: inputs may be masked/unmasked, but supervision always uses the masked versions during training. Results are shown for three test settings: unmasked, masked, and mixed. In the mixed case, only the fixed image is masked by ROI. \textbf{Bolded values} indicate statistically significant improvement over other training configuration under the same architecture~(paired Wilcoxon signed-rank test, $\alpha = 0.01$).
    \label{t:masking}}
    \adjustbox{max width = 0.99 \textwidth}
    {{\renewcommand{\arraystretch}{1.5}
        \begin{tabular}{r r C{0.1\textwidth} r c C{0.12\textwidth} r c C{0.12\textwidth} r c
        }
            \toprule
            & & \multicolumn{2}{c}{\textbf{Test (Masked)}} & \quad & \multicolumn{2}{c}{\textbf{Test (Unmasked)}} & \quad & \multicolumn{2}{c}{\textbf{Test (Mixed)}}
            \\
            \cmidrule(lr){3-4} \cmidrule(lr){6-7} \cmidrule(lr){9-10}
            \textbf{Method} & \textbf{Train} &
            \textbf{TRE} $\bolddownarrow$ & \multicolumn{1}{c}{\textbf{NDV} $\bolddownarrow$} & \quad & \textbf{TRE} $\bolddownarrow$ & \multicolumn{1}{c}{\textbf{NDV} $\bolddownarrow$} & \quad & \textbf{TRE} $\bolddownarrow$ & \multicolumn{1}{c}{\textbf{NDV} $\bolddownarrow$}
            \\
            \cmidrule(lr){1-1} \cmidrule(lr){2-2} \cmidrule(lr){3-4} \cmidrule(lr){6-7} \cmidrule(lr){9-10}
            &
            Masked & $0.080\pm0.023$ & $<0.01\%$ & & $0.130\pm0.036$ & $0.07\%$ & & $0.459\pm0.069$ & $0.38\%$ 
            \\
            \rowcolor{cyan!10} \cellcolor[gray]{1} &
            Unmasked & $0.133\pm0.037$ & $<0.01\%$ & & $0.078\pm0.021$ & $<0.01\%$ & & $0.169\pm0.040$ & $0.01\%$
            \\
            & Random & $0.075\pm0.019$ & $<0.01\%$ & & $0.103\pm0.028$ & $0.03\%$ & & $0.241\pm0.054$ & $0.02\%$
            \\
            \rowcolor[gray]{.90} \cellcolor[gray]{1}
            \multirow{-4}{*}{VoxelMorph\cite{balakrishnan2019tmi}}
            &
            Ours & $0.077\pm0.022$ & $<0.01\%$ & & $0.077\pm0.021$ & $<0.01\%$ & & $\mathbf{0.077\pm0.021}$ & $<0.01\%$
            \\
            % ---------------------------------
            \cmidrule(lr){1-1} \cmidrule(lr){2-2} \cmidrule(lr){3-4} \cmidrule(lr){6-7} \cmidrule(lr){9-10}
            &
            Masked & $0.076\pm0.031$ & $<0.01\%$ & & $0.275\pm0.085$ & $0.06 \%$ & & $1.017\pm0.309$ & $4.55\%$ 
            \\
            \rowcolor{cyan!10} \cellcolor[gray]{1} &
            Unmasked & $0.175\pm0.051$ & $<0.01\%$ & & $0.076\pm0.024$ & $<0.01\%$ & & $0.165\pm0.051$ & $0.02\%$
            \\
            & Random & $0.078\pm0.026$ & $<0.01\%$ & & $0.093\pm0.030$ & $0.01\%$ & & $0.239\pm0.080$ & $0.36\%$
            \\
            \rowcolor[gray]{.90} \cellcolor[gray]{1}
            \multirow{-4}{*}{TransMorph\cite{chen2022mia}}
            &
            Ours & $\mathbf{0.069\pm0.021}$ & $<0.01\%$ & & $\mathbf{0.068\pm0.021}$ & $<0.01\%$ & & $\mathbf{0.066\pm0.020}$ & $<0.01\%$
            \\
            % ---------------------------------
            \cmidrule(lr){1-1} \cmidrule(lr){2-2} \cmidrule(lr){3-4} \cmidrule(lr){6-7} \cmidrule(lr){9-10}
            &
            Masked & $0.053\pm0.015$ & $0.01\%$ & & $0.296\pm0.069$ & $<0.01\%$ & & $0.686\pm0.070$ & $<0.01\%$ 
            \\
            \rowcolor{cyan!10} \cellcolor[gray]{1} &
            Unmasked & $0.200\pm0.060$ & $<0.01\%$& & $0.053\pm0.014$ & $<0.01\%$ & & $0.927\pm0.075$ & $0.01\%$
            \\
            & Random & $0.050\pm0.013$ & $0.01\%$ & & $0.048\pm0.012$ & $<0.01\%$ & & $0.051\pm0.013$ & $<0.01\%$
            \\
            \rowcolor[gray]{.90} \cellcolor[gray]{1}
            \multirow{-4}{*}{VFA\cite{liu2024jmi}}
            &
            Ours & $0.050\pm0.013$ & $0.01\%$ &  & $0.049\pm0.011$ & $<0.01\%$ & & $0.050\pm0.013$ & $0.01\%$
            \\
        \bottomrule
        \end{tabular}
    }}
\end{table*}

\subsection{Lung Mask Agnostic Lung CT Registration}
To demonstrate the use of surrogate supervision in learning ROI mask agnostic registration, we used the NLST dataset~\cite{prorok1994cjo} prepared by the 2022~Learn2Reg challenge for intra-subject inhale-exhale lung CT registration.
The original training set in the challenge was split into $79$ pairs for training, $10$ for validation, and $10$ for testing.
Expert-annotated landmarks within the lung region are available for all pairs, which allows the computation of target registration error~(TRE).
Each scan includes a lung mask.

We train each architecture under four settings:
\begin{enumerate}
    \item Masked: both inputs masked by ROI, 
    \item Unmasked: neither input masked, 
    \item Random: both inputs either masked or unmasked, chosen randomly per sample, 
    \item Ours: inputs may be masked/unmasked, but supervision always learns on the masked versions.
\end{enumerate}
Settings~(1) and~(2) reflect standard practices commonly adopted in the literature, while setting~(3) represents a straightforward extension that could be practically deployed to reduce the model’s reliance on specific ROI masks.
All models were trained using a combination of NCC loss, TRE loss, and diffusion regularizer with weights adopted from previous work~\cite{liu2024jmi}.
After training, we evaluated model performance using TRE on three different settings: unmasked, masked, and mixed. In the mixed case, only the fixed image is masked by ROI.
These settings simulate varying real-world test scenarios where the data preparation process may differ between training and deployment.

Table~\ref{t:masking} reports the results, with bolded values indicating statistically significant TRE improvement over other training configurations under the same architecture~(paired Wilcoxon signed-rank test, $\alpha = 0.01$).
Across all three architectures, we observed that models generally perform well when the training and testing configurations were aligned (\eg~masked model on masked inputs or unmasked model on unmasked inputs), reflecting the fact that these models adapt closely to the specific preprocessing conditions seen during training.
However, substantial performance degradation occurs when training and testing conditions differ.
In contrast, surrogate supervision consistently achieves strong performance across all testing scenarios, matching the accuracy of the masked model on masked inputs and the unmasked model on unmasked inputs. It also maintains consistent accuracy under the mixed setting, despite never being trained explicitly for this case.
This robustness arises because surrogate supervision guides the model to focus on the lung region regardless of input masking, whereas other strategies rely heavily on the form of the training inputs. 
Random masking partially alleviates the mismatch issue by exposing the model to both masked and unmasked samples during training.
This proves highly effective for VFA, even yielding competitive results in the mixed case. Yet, for both VoxelMorph and TransMorph, random masking tends to compromise performance in one or more test scenarios and does not reliably improve the mixed case.

\subsection{Multi-modal Registration Revisit}
Previous work has shown that when paired and pre-aligned multi-modal images are available, the registration task can be reformulated as a mono-modal problem by applying the estimated deformation to a surrogate image in the same modality as the fixed image. This bypasses the challenges of designing multi-modal similarity measures. Since several advanced multi-modal loss functions have been implemented for modern deep learning frameworks in the recent years, we revisit the problem of multi-modal registration to assess how the surrogate supervision approach compares to these multi-modal similarity losses, providing an updated perspective on its relevance and performance in current practice.

We trained models to register T2-weighted~(T2w) MR images to T1-weighted~(T1w) MR images using 400 subjects from the IXI dataset~\cite{IXI2007}.
Training involved inter-subject registration, where a randomly selected T2w image and a T1w image from different subjects served as the moving and fixed inputs, respectively.
The validation and test sets consisted of 40 and 135 fixed image pairs, respectively.
All scans were preprocessed with N4 inhomogeneity correction~\cite{tustison2010tmi}, skull stripping via HD-BET~\cite{isensee2019hbm}, and rigid alignment using Advanced Normalization Tools~(ANTs)~\cite{avants2011reproducible}.
Segmentation maps of 133 anatomical structures were generated using a whole-brain segmentation algorithm~\cite{huo20193d}, and DSC was computed between the warped and fixed labels.

Each architecture was trained with four configurations:
\begin{enumerate}
    \item NCC:~normalized cross-correlation loss,
    \item CR:~correlation ratio loss,
    \item MI:~mutual information loss,
    \item Ours:~surrogate supervision with NCC loss.
\end{enumerate}
For the CR and MI configurations, we tuned the regularization weight $\lambda$ using TransMorph, evaluating values from \{0.5, 1, 2, 5, 10\}.
For both the CR and MI configurations, the best validation accuracy was achieved with $\lambda = 5$, which was then adopted for VoxelMorph and VFA.
When using NCC loss, we directly adopted $\lambda=1$, following prior works on mono-modal brain MR registration.

Table~\ref{t:multi-modal} summarizes the results, with bolded values indicating statistically significant DSC improvement over the other comparison methods of the same architecture (paired Wilcoxon signed-rank test, $\alpha = 0.01$)
We observe that the NCC loss without surrogate supervision consistently yields the lowest performance across all three architectures, which was expected given that NCC is not designed for multi-modal registration where intensity profiles of the images differ between modalities.
Across the different architectures and training configurations, surrogate supervision with NCC achieved performance on par with mutual information when used in VoxelMorph, and yielded statistically significant improvements over baseline similarity losses in all the other settings.
Due to our choice of $\lambda$ values, training with NCC loss produces larger non-diffeomorphic volumes.
However, lowering $\lambda$ for CR and MI does not improve their accuracy on the validation set.
Our results indicates that when paired scans are available, surrogate supervision is not only competitive with carefully designed multi-modal similarity measures, but can also provide clear advantages in practice. Despite the availability of modern multi-modal loss functions, reformulating the task through surrogate supervision remains a strong and often superior alternative.

\begin{table}[!b]
    \centering
    \caption{Quantitative results for multi-modal T2-weighted to T1-weighted MR registration on the IXI dataset. We report mean Dice Similarity Coefficient~(DSC) on $133$ anatomical label overlaps for VoxelMorph, TransMorph, and VFA architectures, each trained with normalized cross correlation~(NCC) loss, Correlation ratio~(CR) loss, mutual information~(MI) loss, or surrogate supervision with NCC loss~(Ours). We also report non-diffeomorphic volume~(NDV)~\cite{liu2022finite} for each configuration and architecture pair. \textbf{Bolded values} indicate statistically significant DSC improvement over the other comparison methods of the same architecture (paired Wilcoxon signed-rank test, $\alpha = 0.01$).
    \label{t:multi-modal}}
    \adjustbox{max width = 0.99 \textwidth}
    {{\renewcommand{\arraystretch}{1.5}
        \begin{tabular}{r r C{0.1\textwidth} r c
        }
            \toprule
            \textbf{Method} & \textbf{Train} &
            \textbf{DSC} $\bolduparrow$ & \multicolumn{1}{c}{\textbf{NDV} $\bolddownarrow$} &
            \\
            \cmidrule(lr){1-2}\cmidrule(lr){3-5}
            & NCC & $0.540\pm0.089$ & $0.47\%$ &
            \\
            \rowcolor{cyan!10} \cellcolor[gray]{1} & CR\cite{chen2025correlation} & $0.597\pm0.070$ & $<0.01\%$ &
            \\
            & MI\cite{guo2019multi} & $0.625\pm0.062$ & $<0.01\%$&
            \\
            \rowcolor[gray]{.90} \cellcolor[gray]{1}
            \multirow{-4}{*}{VoxelMorph\cite{balakrishnan2019tmi}}
            &
            Ours & $0.628\pm0.081$ & $0.49\%$&
            \\
            \cmidrule(lr){1-2}\cmidrule(lr){3-5}
             & NCC & $0.603\pm0.081$ & $0.36\%$ &
            \\
            \rowcolor{cyan!10} \cellcolor[gray]{1} & CR\cite{chen2025correlation} & $0.644\pm0.055$ & $0.01\%$ &
            \\
            &
            MI\cite{guo2019multi} & $0.672\pm0.058$ & $<0.01\%$&
            \\
            \rowcolor[gray]{.90} \cellcolor[gray]{1}
            \multirow{-4}{*}{TransMorph\cite{chen2022mia}}
            &
            Ours & $\mathbf{0.721\pm0.050}$ & $0.36\%$&
            \\
            \cmidrule(lr){1-2} \cmidrule(lr){3-5}
            & NCC & $0.612\pm0.040$ & $<0.01\%$ &
            \\
            \rowcolor{cyan!10} \cellcolor[gray]{1} & CR\cite{chen2025correlation} & $0.678\pm0.027$ & $<0.01\%$ &
            \\
            &
            MI\cite{guo2019multi} & $0.726\pm0.022$ & $<0.01\%$&
            \\
            \rowcolor[gray]{.90} \cellcolor[gray]{1}
            \multirow{-4}{*}{VFA\cite{liu2024jmi}}
            &
            Ours & $\mathbf{0.776\pm0.018}$ & $0.03\%$&
            \\
        \bottomrule
        \end{tabular}
    }}
\end{table}

\section{Discussion}

Surrogate supervision can be interpreted as a general strategy for defining loss functions. In registration and related tasks, designing effective similarity measures is often difficult.
While task-specific surrogate signals have been used in isolated cases, framing surrogate supervision as a unified principle reveals its broader potential.
The key idea is that, instead of relying on increasingly complex loss formulations, one can identify or construct appropriate surrogate representations and then apply simple, well-established loss functions in the surrogate domain to achieve sophisticated training objectives.
This reframing offers a perspective shift, providing a systematic way to rethink supervision design.

Surrogate supervision may appear similar to data augmentation because both involve additional data derived from the originals. Data augmentation operates on the input domain by introducing variability through perturbations or transformations (\textit{e.g.}, intensity and geometric changes).
By contrast, surrogate supervision operates on the supervision domain: it redefines the target of comparison.
By modifying the supervision domain, surrogate supervision can embed domain knowledge directly into the training signal and support optimization with simple, well-established losses even on heterogeneous data.
The two strategies are therefore complementary and can be combined naturally, as demonstrated in our mask-agnostic registration experiment.

Compared to approaches that focus on scaling up training data in the input domain, surrogate supervision highlights the importance of data quality. Specifically, the construction or selection of surrogate targets that provide meaningful and reliable supervision signals.
This introduces a need for domain knowledge and thoughtful assumptions about which aspects of the data are most trustworthy or clinically relevant.
While this requirement could be seen as a limitation, it also reflects a broader shift: as models grow in capacity and generality, the quality and structure of training data become just as critical as dataset size or model complexity~\cite{wang2022self}.

\section{Conclusion}

We introduced surrogate supervision, a general training paradigm that unifies and extends prior works for robust and generalizable deep learning–based deformable image registration. The key insight behind this approach is that supervision can be applied in a surrogate domain, rather than the raw input image domain. By leveraging appropriate surrogates, models can learn robustly from heterogeneous data using standard similarity losses or incorporate domain knowledge from existing preprocessing pipelines, all without introducing additional complexity at deployment.

Importantly, the effectiveness of surrogate supervision depends primarily on the availability and quality of surrogate representations, rather than on architectural or loss-design choices.
This flexibility allows surrogate supervision to extend naturally beyond the cases we demonstrated in this paper.
For example, hybrid imaging~(\textit{e.g.}, nuclear medicine/CT) could benefit from surrogate supervision, where anatomical information guides the training of functional modalities without being required at test time.
In addition to artifact correction, surrogate supervision can also integrate processes such as super-resolution~\cite{zhao2019applications} or image harmonization~\cite{krishnan2024lung,zuo2023haca3}, effectively allowing a registration model to serve as a universal input interface.

\bibliographystyle{unsrt}
\bibliography{references}

\end{document}